\documentclass[pmlr]{jmlr}
\usepackage{colortbl}

\makeatletter
\def\set@curr@file#1{\def\@curr@file{#1}} 
\makeatother
\usepackage{siunitx} 
\usepackage[utf8]{inputenc} 
\usepackage[T1]{fontenc}    
\usepackage{hyperref}       
\usepackage{url}            
\usepackage{booktabs}       
\usepackage{amsfonts}       
\usepackage{nicefrac}       
\usepackage{microtype}      
\usepackage{booktabs}
\usepackage{caption}
\usepackage{multirow} 
\usepackage{rotating}
\usepackage{contour}
\usepackage{xcolor}
\usepackage{amsmath}
\usepackage{textcomp}
\usepackage{enumitem}
\usepackage{wrapfig}
\usepackage{lmodern}
\usepackage{natbib}
\usepackage{graphicx}
\DeclareGraphicsExtensions{.pdf,.png,.jpg}  

\newcommand{\etal}{et~al.}

\theorembodyfont{\upshape}
\theoremheaderfont{\scshape}
\theorempostheader{:}
\theoremsep{\newline}

\jmlrvolume{252}
\jmlryear{2024}
\jmlrworkshop{Machine Learning for Healthcare}

\usepackage{xcolor}
\usepackage[normalem]{ulem}

\usepackage[T1]{fontenc}
\usepackage{url}

\title[\color{red}\uwave{\color{black}\textbf{\textit{MedAutoCorrect}}}]{\color{red}\uwave{\color{black}\textbf{\textit{MedAutoCorrect}}}\color{black}\\ Image-Conditioned Autocorrection in Medical Reporting
  }

\author{\Name{Arnold Caleb Asiimwe}
       \Email{a.asiimwe@columbia.edu}\\ 
       \addr Columbia University \\
       New York, NY, United States
       \AND
       \Name{D\'{i}dac Sur\'{i}s}
       \Email{didac.suris@columbia.edu}\\ 
       \addr Columbia University \\ 
       New York, NY, United States
       \AND
       \Name{Pranav Rajpurkar}
       \Email{pranav\_rajpurkar@hms.harvard.edu}\\ 
       \addr Harvard Medical School \\
       Cambridge, MA, United States
       \AND
       \Name{Carl Vondrick}
       \Email{cvondrick@cs.columbia.edu}\\ 
       \addr Columbia University \\
       New York, NY, United States
       }

\begin{document}

\maketitle

\maketitle
\begin{abstract}
In medical reporting, the accuracy of radiological reports, whether generated by humans or machine learning algorithms, is critical. We tackle a new task in this paper: image-conditioned autocorrection of inaccuracies within these reports. Using the MIMIC-CXR dataset, we first intentionally introduce a diverse range of errors into reports. Subsequently, we propose a two-stage framework capable of pinpointing these errors and then making corrections, simulating an \textit{autocorrection} process. This method aims to address the shortcomings of existing automated medical reporting systems, like factual errors and incorrect conclusions, enhancing report reliability in vital healthcare applications. Importantly, our approach could serve as a guardrail, ensuring the accuracy and trustworthiness of automated report generation. Experiments on established datasets and state of the art report generation models validate this method's potential in correcting medical reporting errors.   

\end{abstract}
    
\section{Introduction}
\label{sec:intro}

Medical reports, particularly in radiology, are cornerstones of healthcare \citep{Brady2018Radiology}. They offer crucial interpretations of medical images that directly affect clinical decision-making and patient care \citep{ ESR2011GoodPractice}. Given their role, ensuring the accuracy and dependability of these reports is vital.

There has been a surge in efforts to automate the generation of medical reports \citep{nguyen-etal-2021-automated}. These automated approaches offer the promise of uniformity and the potential to reduce the heavy workload faced by radiologists. Yet, reports created by both humans and radiology report generation systems are susceptible to errors \citep{jing-etal-2018-automatic}. For humans, issues such as fatigue and high case volumes can lead to mistakes \citep{Brady2017ErrorDiscrepancy}. Evidence gathered during the plain film era suggested a radiologist error rate of around 3–5\% in daily practice \citep{maskell2019error}. For radiology report generation systems, inaccuracies can arise from limited data, built-in biases, or model constraints, resulting in errors such as incorrect predictions or omissions of findings, misidentification of their location or severity, inappropriate comparative references, or failure to note changes from previous studies \citep{miura-etal-2021-improving, Yu2023Evaluating}. If integrated into clinical practice, these inaccuracies could have profound consequences for patient care.
\vspace{-0.4cm}
\begin{wrapfigure}{r}{0.5\textwidth}
\includegraphics[width=0.5\textwidth]{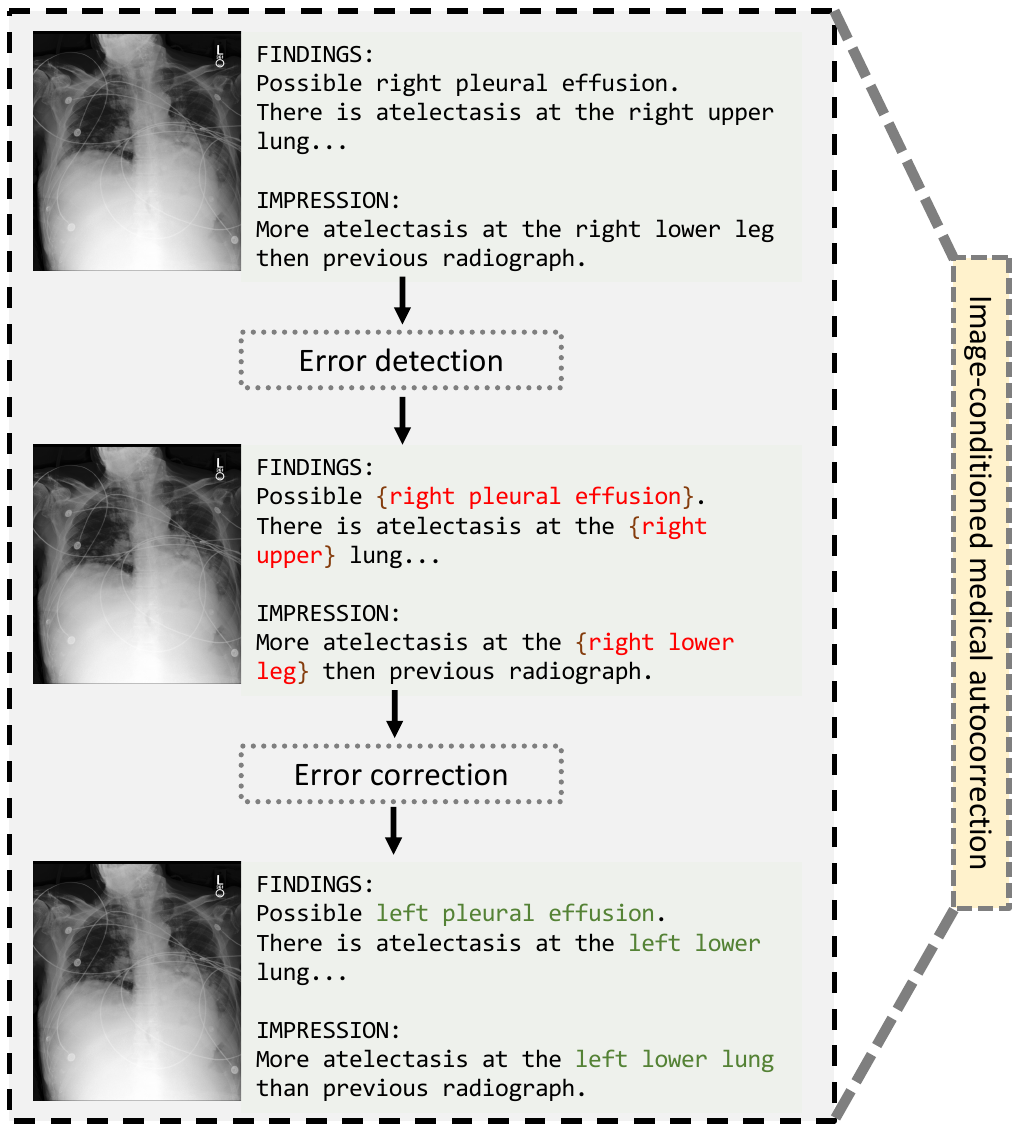}
\caption{Overview of our \texttt{DETECT + CORRECT} error-correction method}
\label{fig:overview}
\vspace{-0.5cm}
\end{wrapfigure}

In this paper, we propose an approach that uses visual information to detect and auto-correct errors in medical reports. Recently, there have been very successful methods for aligning images and language together \citep{radford2019language}, which we adopt for the purpose of error detection and correction. However, due to the significant distribution shift, models trained on Internet data are not directly applicable to the medical domain. Additionally, it is challenging to collect a large paired training set of erroneous reports and their corrected counterparts. We introduce a procedure to synthetically inject errors into correct medical reports, allowing us to consequently learn to detect and remove them. We develop and compare several conditioning mechanisms to detect errors and make corrections. Our approach, illustrated in Figure~\ref{fig:overview}, can identify and rectify errors whether they originate from machine learning models or human radiologists.

\subsection*{Generalizable Insights about Machine Learning in the Context of Healthcare}
Our findings show that incorporating autocorrection into radiological report generative models can significantly enhance the natural language generation (NLG) scores (Table~\ref{tab:second_eval}) of retrieval-based models. This improvement is achieved even on models that are not initially optimized for state-of-the-art (SOTA) performance. By precisely addressing the errors introduced by these generative models, our approach elevates their outputs to SOTA levels. Additionally, we provide qualitative evidence by showcasing specific instances where our model successfully corrects errors in generated reports, acting as a guardrail (Fig~\ref{fig:retrieval_example}, Sec~\ref{sec:retrieval}). We hope the community will find our tasks, datasets (subject to appropriate permissions), and models useful towards applications of computer vision in healthcare.

\section{Related Work}
\label{sec:related_work}

Autocorrection has become an indispensable tool in human-computer interfaces, such as Microsoft Word or Google Docs.
The main goal of these tools has been to uphold both syntactic and semantic integrity within written documents. We recontextualize the concept of autocorrection to pertain specifically to the accuracy of radiological reports grounded in the images they describe. Our focus is to mitigate factual discrepancies that may arise from the oversight of medical professionals and in the interpretations made by radiology report generative models.

\subsection{Radiology Report Generation}

Recent efforts in radiology report generation have predominantly focused on enhancing model accuracy through architectural design changes, often neglecting the correction of errors in generated reports \citep{Tanida_2023, chen2022crossmodal, miura-etal-2021-improving, liu2019clinically}. Initially structured as an image-captioning task \citep{wang2018tienet, yuan2019automatic}, these models have evolved significantly. For instance, region-guided approaches, like those in \citep{Tanida_2023}, improve performance by generating reports for specific anatomical regions in chest X-rays. While early models primarily employed CNN-RNN or stacked CNN-LSTM architectures \citep{chen2020say, nguyen-etal-2021-automated, 10.5555/3326943.3327084, wang2018tienet, yuan2019automatic}, recent studies have pivoted towards Transformer and attention-based encoders for enhanced cross-modal interaction \citep{NIPS2017_3f5ee243}. Others include relation memory units and memory matrices \citep{chen2022crossmodal}, systems for contrasting normal and abnormal images \citep{10.5555/3326943.3327084}, and the integration of medical knowledge graphs \citep{10.1609/aaai.v33i01.33018650}.

Radiology report generation methods are largely inspired by image captioning techniques in computer vision \citep{ALFARGHALY2021100557, chen2022crossmodal, chen-etal-2020-generating, 10.1609/aaai.v33i01.33018650, 10.5555/3326943.3327084, 9733170, 10.1007/978-3-031-16452-1_54, cornia2020meshedmemory, sanh2020distilbert, vedantam2015cider, wang2018tienet}. Despite parallels with general image captioning, radiology report generation faces unique challenges: they are typically more detailed and diverse, covering multiple anatomical regions. Additionally, the need to accurately describe specific abnormalities is complicated by data biases towards standard images and reports, leading to the generation of erroneous reports \citep{chen-etal-2020-generating}. To counter this, some methods have adopted object detection strategies from the dense image captioning domain \citep{johnson2015densecap, 10.1609/aaai.v33i01.33018650, 9733170, yin2019context}, aiming to both localize and describe individual salient regions in images, typically by conditioning a language model on specific region features. These developments represent various approaches undertaken to enhance the accuracy of radiological reports.

\subsection{The Gap}
Despite the progress in both autocorrection and radiological report generation, a conspicuous gap remains. In radiological report generation, the models are still very error prone \citep{chen2022crossmodal} and research into the mitigation of the created errors has been minimal, with a focus on mainly architectural changes to enhance report accuracy. Text-based autocorrection systems can only take us as far as correcting grammatical and syntactical discrepancies in text. In this paper, we introduce an image-conditioned autocorrection framework, not only identifying a critical need within the current landscape of radiological report generation but also offering a targeted solution. This method has the potential to assist radiologists by identifying and rectifying inaccuracies within their radiological reports.

\section{Methods}
\label{sec:methods}

Our approach consists of three distinct steps. First, in Sec.~\ref{sec:error_injection} we propose a procedure to obtain a dataset of medical reports with annotated errors. Second, in Sec.~\ref{sec:detection} we explain how we can use this dataset in order to train a model that detects such errors. Third, in Sec.~\ref{sec:correction} we build on top of the error identification module and propose an error correction module. Finally, we provide more details on training and inference procedures in Secs.~\ref{sec:training} and \ref{sec:inference}.

\begin{figure*}[htbp]
\centerline{\includegraphics[width=0.85\textwidth, height=188pt]{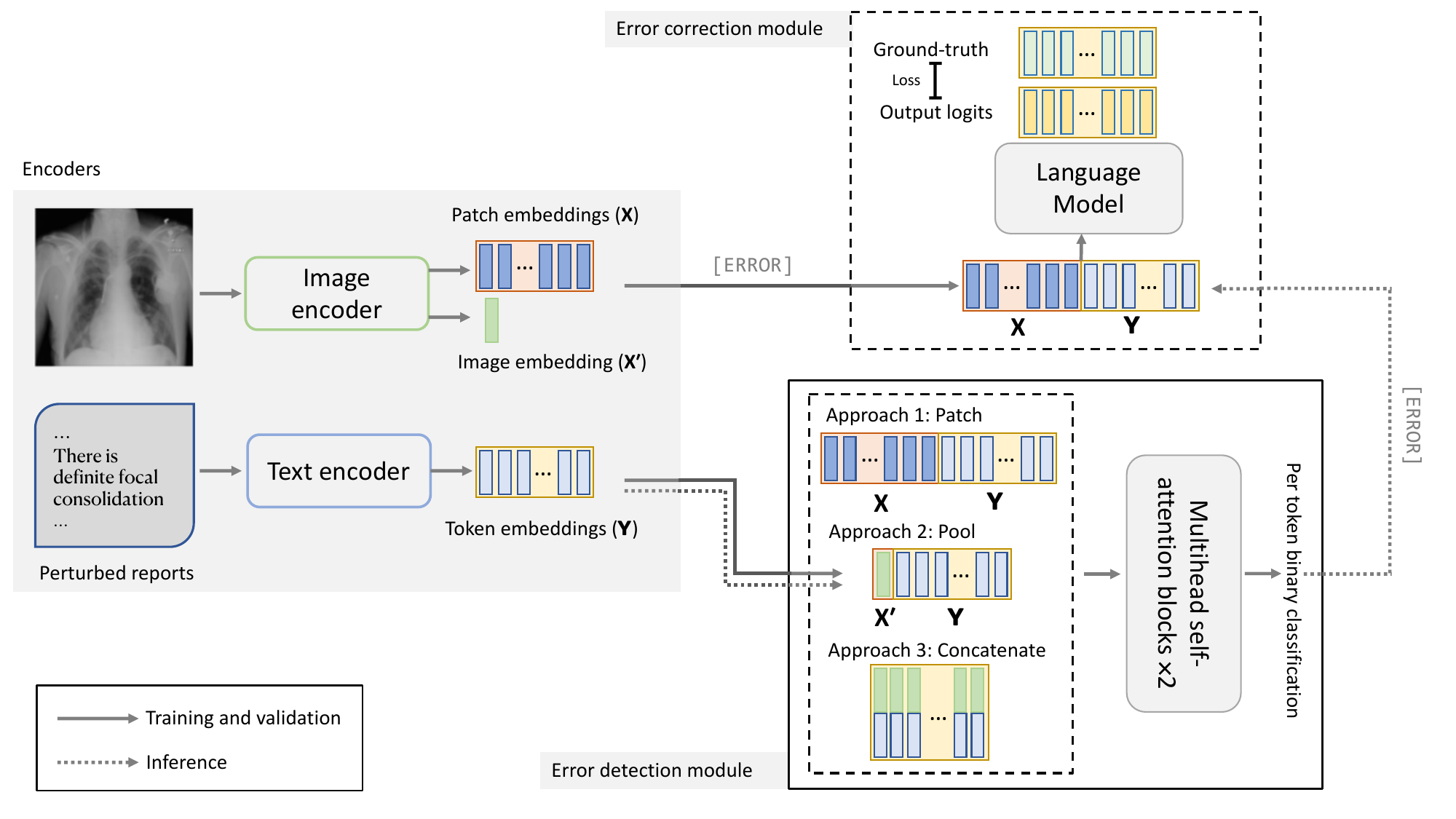}}
\caption{\textbf{Overview of the Proposed Framework for Autocorrecting Radiology Reports}. The training phase initiates with separate encoding processes for images and text. The encoded representations are then processed by an error identification module, which utilizes three distinct approaches to detect inaccuracies. Subsequently, a language model is fine-tuned on the image-contextualized reports, where injected errors are represented by \texttt{[ERROR]} tokens. During inference, the model applies the error detection mechanism to localize errors that later are replaced with the masked tokens which are corrected by the error correction mechanism.}
\vspace{-0.5cm}
\label{fig:architecture}
\end{figure*}

\subsection{Error injection}
\label{sec:error_injection}
\begin{wrapfigure}{r}{0.4\textwidth}
\vspace{-0.8cm}
\centerline{\includegraphics[width=0.4\textwidth]{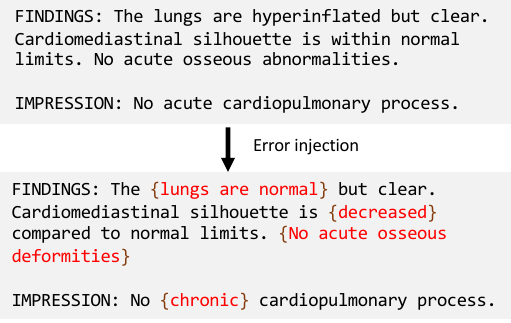}}
\caption{\textbf{Error injection example}. In this example, we automatically introduce errors that fall within the categories of incorrect prediction.}
\vspace{-0.5cm}
\label{fig:error_injection}
\end{wrapfigure}
To simulate common errors found in radiological reporting, we introduce specific inaccuracies into the reports from the MIMIC-CXR dataset \citep{Johnson2019MIMIC}, which consists of chest X-ray images paired with free-text radiological reports. These intentionally induced errors are categorized into six major types as identified in \citep{YU2023100802}: 1) False prediction of findings, 2) Incorrect location/position of findings, 3) Incorrect severity of findings, 4) Mention of comparisons not present in the reference impression, 5) Omission of findings, and 6) Omission of comparison describing a change from a previous radiological image. In this study, we concentrate on the first four categories, providing detailed explanations next.

1) A false prediction of findings occurs when the report incorrectly identifies a medical condition of finding that is not present in the radiological images. 2) Incorrect location/position of findings involves identifying the right finding but attributing it to the wrong anatomical location within the image. 3) Incorrect severity of findings arises when the report either underestimates or overestimates the seriousness of a condition evident in the radiological images. 4) Mention of comparison not present in the reference impression refers to instances where the report includes comparative references to previous images that are not  part of the reference impression.

\begin{figure*}[htbp] 
\centerline{\includegraphics[width=1.\textwidth]{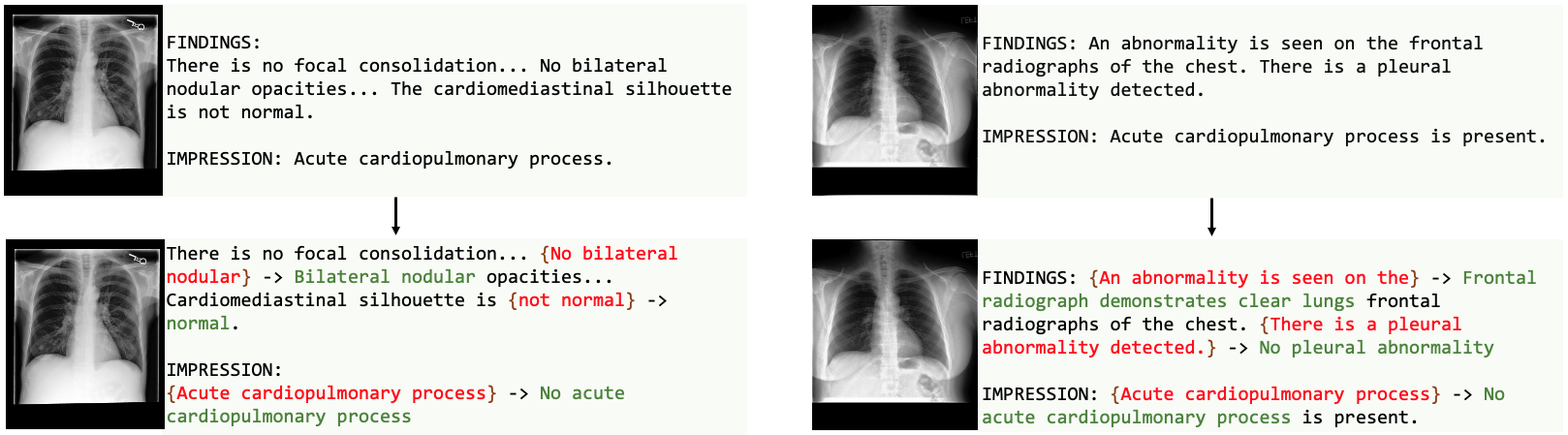}}
\caption{\textbf{Qualitative example of medical autocorrection:} The top section shows the initial report with errors. The bottom section displays the report after processing by our model, with corrections in green and erroneous terms struck through. This exemplifies the model's capability to identify and rectify inaccuracies within clinical text.}
\label{fig:q_4}
\end{figure*}

The error injection process involves the \textbf{intentional} systematic introduction of errors into the MIMIC-CXR radiological reports (see Figure~\ref{fig:error_injection} and \ref{fig:q_4}). This process is executed through a combination of manual techniques and automated injection, explained next, together contributing to 120,123 erroneous reports. For each report, two erroneous versions of the same report are created and contain a combination of error error types mentioned above.

\textbf{Automated Error Injection}. We automatically introduce errors using a generative large language model (LLM). This allows us to generate more diverse examples that do not fit into specific patterns and follow a natural language distribution. We provide the LLM with specific prompts crafted to reflect the nuances of radiological reporting. Each prompt contains a segment of a real MIMIC-CXR report, followed by instructions to alter it in a way that mimics one or a combination of the four identified error categories mentioned. Specifically, we use OpenAI's GPT-4 \citep{openai2023gpt4}. 

To illustrate, for the category of incorrect predictions, a typical prompt to the LLM would read: ``Given the following excerpt from a radiological report, modify it to contain an incorrect diagnostic prediction while maintaining a plausible and professional tone.'' A similar approach was used for the other categories, ensuring that each generated error is contextual and aligned with real-world reporting scenarios. 

\begin{figure*}
\vspace{-0.5cm}
\centerline{\includegraphics[width=1.1\textwidth]{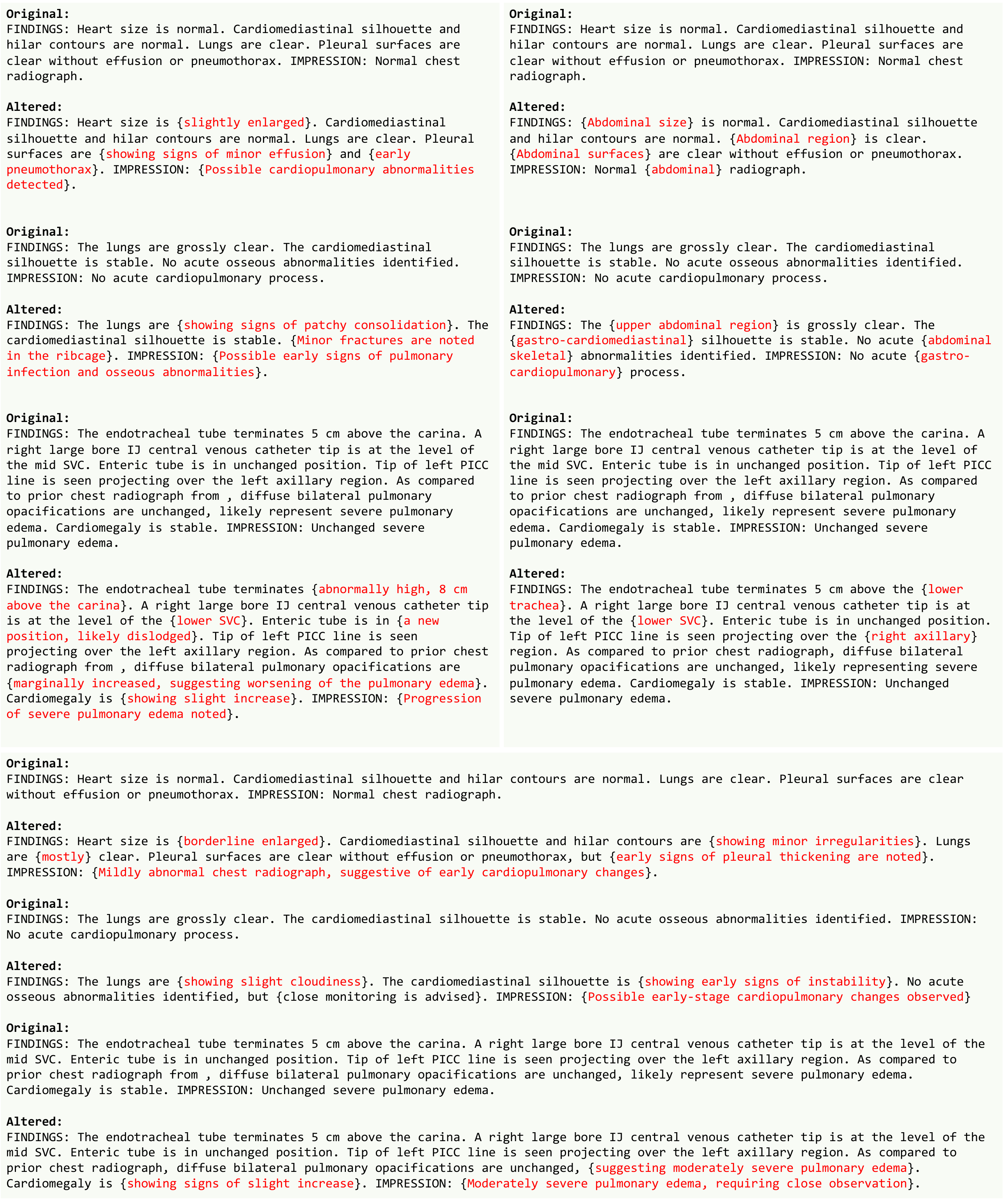}}
\caption{Overview of varied error types in radiological reports, as altered via GPT-4 prompts. Top-left focuses on false predictions, top-right on mislocations, and the bottom on severity misjudgments, illustrating common error types in clinical radiology and their potential impacts on diagnostic accuracy. Some reports in the dataset consist of a mixture of errors.}
\label{fig:main_all}
\end{figure*}

\textbf{Manual Error Injection}. We manually alter location information, medical condition labels, and the severity of such conditions. See Figure ~\ref{fig:main_all} for some examples.

\subsection{Error Detection Module}
\label{sec:detection}

Our approach implements error detection as a per-token classification task (see bottom right of Figure~\ref{fig:architecture}). Within this framework, each token of a given erroneous radiological report, accompanied by an image, is subject to binary classification. This process discerns the correctness of each token within the image's contextual parameters. 

We use an image encoder -- specifically, a Vision Transformer \citep{dosovitskiy2021image} -- that has been fine-tuned on chest X-ray images \citep{irvin2019chexpert}. We use the encoder to create two distinct forms of embeddings: patch embeddings and a singular, attention-pooled image embedding. These embeddings are subsequently projected onto a uniform dimensional space, aligned with that of the token embeddings.

The erroneous radiological report is encoded using GatorTron-medium, a language model optimized for electronic health records~\citep{Yang2022LargeLanguageEHR}, from which token embeddings are extracted. Both the image and text encoders are fixed.

We investigate three image-conditioning strategies. Figure~\ref{fig:architecture} shows an overview:
\begin{enumerate}[wide, labelwidth=!,label=\textbf{\arabic*})]
    \item \textbf{Approach 1: Patch}. We condition on patch embeddings, appending them to the token embeddings:
    \begin{equation}
    \vspace{-0.1cm}
    S_{1} = [P; T],
    \end{equation}
    resulting in a length of \( P + N \), with \( P \) and \( N \) representing the counts of patch and token embeddings, respectively.
    
    \item \textbf{Approach 2: Pool}. We condition on the pooled image embedding by appending it to the token embeddings, yielding a sequence:
    \begin{equation}
    \vspace{-0.1cm}
    S_{2} = [I; T],
    \end{equation}
    with a length of \( 1 + N \).
    
    \item \textbf{Approach 3: Concatenate}. We concatenate the pooled image embedding with each token embedding, producing a sequence of length \( N \) but with a dimensionality of \( 2 \times \text{dim}\_\text{size} \):
    \begin{equation}
    \vspace{-0.1cm}
    S_{3} = \text{concat}(I, T_i) \quad \forall i \in \{1, \ldots, N\},
    \end{equation}
    where \( T_i \) is the \( i \)-th token embedding.
\end{enumerate}

The sequences from each approach are then processed through two multihead self-attention blocks, each with eight heads, followed by a per-token classification.
To handle the imbalance in the classification of tokens (as the number of erroneous tokens is much less than the correct tokens), we employ Focal Loss \citep{lin2018focal} for binary classification, defined as:
\vspace{-0.2cm}
\begin{equation}
    \begin{aligned}
        \mathcal{L}_{\text{detection}} &= -\frac{1}{NM} \sum_{j}^{M} \sum_{i}^{N} \Big[ \alpha_{ji} (1 - p_{ji})^\gamma y_{ji} \log(p_{ji}) + (1 - \alpha_{ji}) p_{ji}^\gamma (1 - y_{ji}) \log(1 - p_{ji}) \Big],
    \end{aligned}
\end{equation}
where \( j \) is the index of the individual report examples, \( i \) is the index of each token within a single report, \( \alpha_{ji} \) is a weighting factor to balance the importance of positive/negative examples, \( p_{ji} \) is the predicted probability of the \( i \)-th token in the \( j \)-th report being corrected, \( y_{ji} \) is the true label, and \( \gamma \) is the focusing parameter that adjusts the rate at which easy examples are down-weighted. \( N \) represents the number of tokens being classified and \( M \) is the number of examples in each batch. We choose a $\gamma$ factor of 2 and set $\alpha$ to 0.85, batch size of 64 and trained on the first 200 tokens (See figure~\ref{fig:distribution} for justification of this choice)of reports for 20 epochs. See suplementary material for details.

\subsection{Error Correction Module}
\label{sec:correction}
Given the detected errors in the radiological reports, the purpose of the error correction module is to rectify them (see top-right of Figure~\ref{fig:architecture}).
We condition the 344M-parameter model GPT-2 Medium \citep{radford2019language}, fine-tuned on a corpus of PubMed abstracts on the images. For this part, the image encoder and the GPT-2 model are not frozen. GPT-2 is an autoregressive neural network that leverages self-attention, conditioning the generation of each token in a sequence on the preceding tokens. The self-attention mechanism can be represented as:
\begin{equation}
SA(Y) = \text{softmax}((Y W_q)(Y W_k)^T)(Y W_v),
\end{equation}
where \(Y\) denotes the token embeddings, and \(W_q, W_k, W_v\) are projection matrices for queries, keys, and values, respectively.

To integrate the image-level patch features with the textual data, we concatenate the patch embeddings \(P\) from the image encoder with the token embeddings \(T\) from the text encoder:
\begin{equation}
S = [P; T],
\end{equation}
where \([;]\) signifies the appending operation, resulting in an extended sequence \(S\) length that combines both modalities.

\begin{figure*}[htbp] 
\centerline{\includegraphics[width=1.\textwidth]{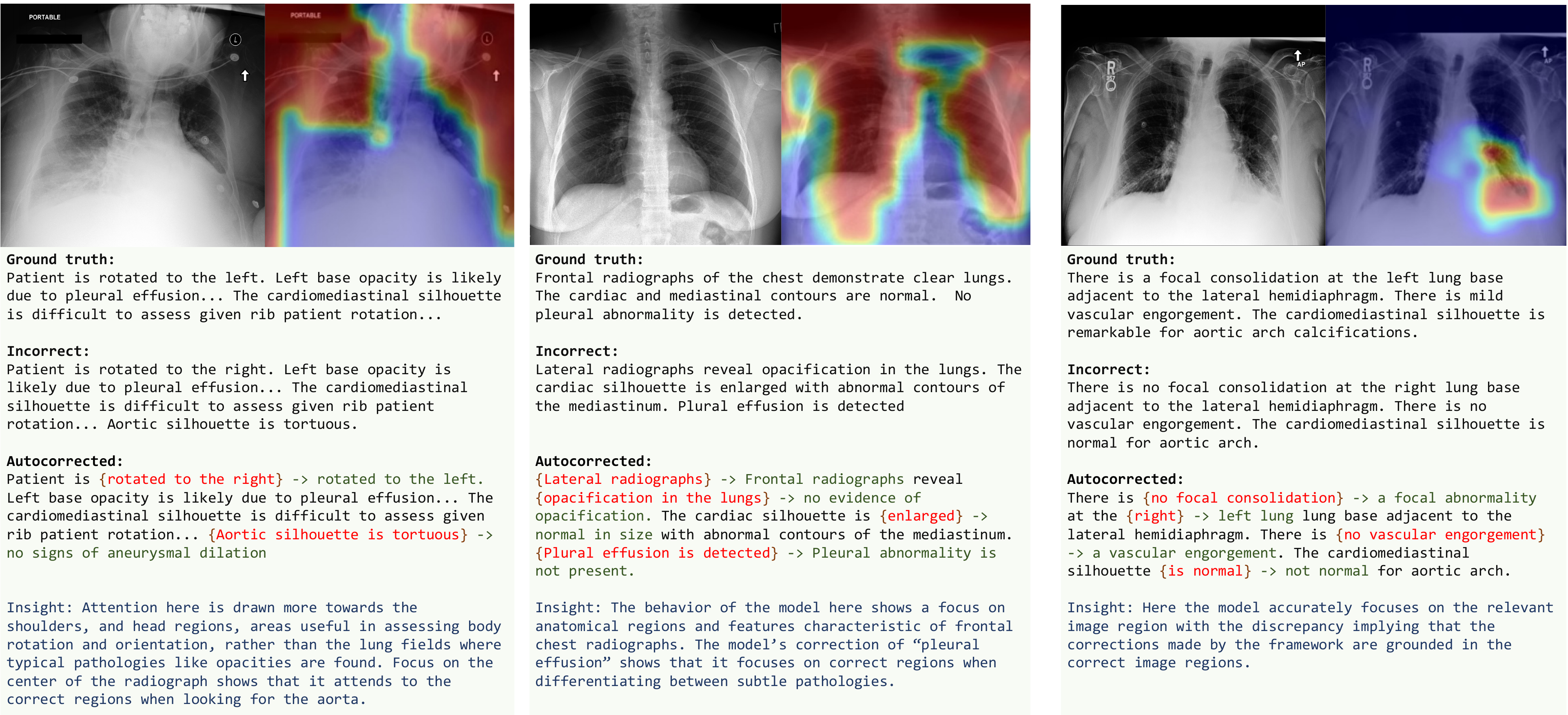}}
\caption{\textbf{Analysing visual attention and autocorrection:} This figure shows how the model's attention is distributed across different regions during error correction. The first image correction focuses on body orientation and the assessment of the aorta, the second on distinguishing between frontal and lateral chest radiographs and the presence of pleural effusion, while the third focuses on the accurate identification of lung abnormalities and aortic arch conditions. We use bicubic interpolation to visualize the attention maps.}
\vspace{-0.5cm}
\label{fig:attention_maps}
\end{figure*}

\begin{figure*}[htbp] 

\centerline{\includegraphics[width=1.\textwidth]{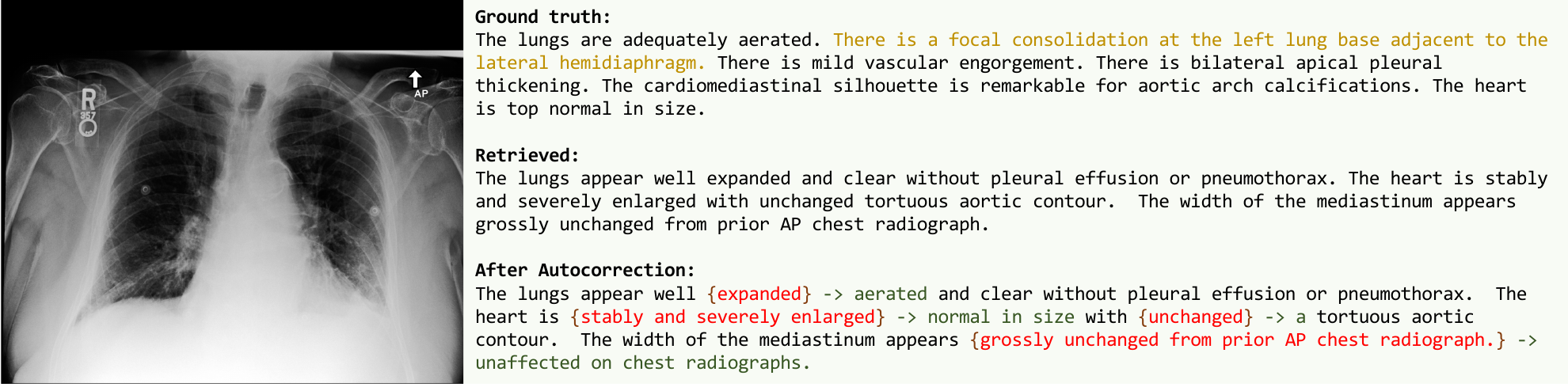}}
\caption{{\textbf{Enhancing Retrieval with Autocorrection:} This illustration shows the \texttt{DETECT+CORRECT} framework in action, where it detects and corrects errors in a report generated by a CLIP-based radiology report generation model (not optimized for SOTA). Highlighted in yellow are sections beyond the scope of correction, as they were not retrieved. Addressing these would shift the task towards report generation, beyond the intended scope of mere correction. In this case, the performance of the autocorrection is only as good as the retrieval model.}}
\vspace{-0.5cm}
\label{fig:retrieval_example}

\end{figure*}

In the fine-tuning process, we utilize a \textbf{masked training strategy} where tokens identified as errors within the text are replaced with the \texttt{[ERROR]} token. This targeted substitution sharpens the learning objective by focusing the model on predicting accurate replacements exclusively at the flagged error positions. As a result, the loss function is calculated solely at these specific locations, directing the model’s learning efforts toward correcting these errors. The correction loss function is defined using cross-entropy, which is calculated between the model’s output logits and the true token embeddings. We adjust for differences in token lengths between the predicted output and the ground truth by padding to match the longer sequence. The formal expression for the correction loss function is:

\begin{equation}
\mathcal{L}_{\text{correction}} = -\frac{1}{MN_{detection}} \sum_{j}^{M} \sum_{i}^{N} \mathbb{1}_{\{i \in \text{error}\}} y_{ji} \log(\hat{y}_{ji}),
\end{equation}
where:

$M$ is the number of examples in each batch. $N_{detection}$ is the number of erroneous tokens being corrected. $j$ is the index of the individual report examples. $i$ is the index of each token within a single report.
$\mathbb{1}_{{i \in \text{error}}}$ is an indicator function that is set to 1 for the positions of the \texttt{[ERROR]} tokens and 0 elsewhere.
$y_{ji}$ is the ground truth for the $i$-th token in the $j$-th report.
$\hat{y}_{ji}$ is the predicted probability for the $i$-th token in the $j$-th report. This ensures that the loss calculated focusing exclusively at locations where the model's predictions for the error tokens need to align with the actual intended medical terminology. 

\subsection{Training}
\label{sec:training}
The training procedure is separated into two distinct phases to create an element of model interpretability. Initially, we focus on the error detection model, which is trained to recognize inaccuracies within reports using the three conditioning strategies outlined. Subsequently, the error correction module is independently trained to generate corrections for the identified and masked errors. This sequential training, as opposed to a joint training approach, allows for a clearer understanding of the error identification and correction mechanisms, ensuring that each proposed correction can be traced back to a specific detected error. We use the AdamW optimizer~\citep{loshchilov2019decoupled} with a batch size of 64 and a learning rate of \(3 \times 10^{-4}\), which is adjusted using a cosine annealing scheduler~\citep{loshchilov2017sgdr}.

\subsection{Inference}
\label{sec:inference}
During inference, we input an image and its corresponding report, which may contain errors, into the error detection module. This module identifies and marks the specific parts of the report that are erroneous. The marked segments are masked with the \texttt{[ERROR]} token. Following this, the masked sections are fed into the autocorrection module, which then makes the necessary corrections to the text, effectively simulating the \textit{autocorrection} process. To accommodate corrections that may vary in length we use beam search~\citep{Freitag_2017} and nucleus sampling~\citep{holtzman2020curious}. 

\section{Experiment Setup}
\label{sec:experiments}

We evaluate our framework across three distinct metrics: the accuracy of error detection within the manipulated reports, the alignment of the autocorrected reports with the original ground truth, and the effectiveness of the model in rectifying errors introduced by a retrieval-based report generation system that follows Endo \etal~\citep{pmlr-v158-endo21a}.

\subsection{Dataset and preprocessing}

\begin{wrapfigure}{r}{0.5\textwidth}
\vspace{-1.4cm}
\centerline{\includegraphics[width=0.5\textwidth]{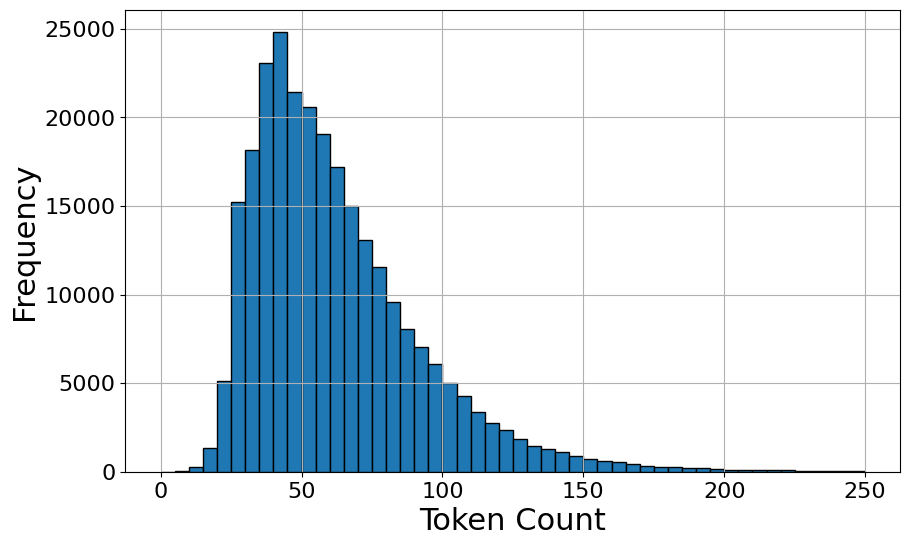}}
\caption{Distribution of report lengths in the MIMIC-CXR dataset that motivated the choice of training on the first 200 tokens. Lengths greater than 250 are not included.}
\vspace{-0.35cm}
\label{fig:distribution}
\end{wrapfigure}

We use the X-rays and the altered reports from the MIMIC-CXR dataset \citep{Johnson2019MIMIC} to train and evaluate our the error detection module, while the original, unaltered reports are reserved for error correction training and evaluation purposes. We adopt an 80:20 split for training and validation, respectively, and assess the model's performance using a separate set of 6,000 image/erroneous text reports reserved for testing.

All images are resized to 224x224 pixels, ensuring the original aspect ratio is maintained, with padding applied as necessary, and they are normalized to have a zero mean and unit standard deviation. Image data augmentation during training includes color jitter, Gaussian noise, and affine transformations following Tanida \etal\citep{Tanida_2023}. Text preprocessing involves retaining only the findings and impressions sections of the reports and removing extra whitespace, such as line breaks. Training, however, is limited to the first 200 tokens (See Figure~\ref{fig:distribution}), padding shorter reports and truncating longer ones. The findings section typically encapsulates the radiologist's observations, while the impressions section provides a concise summary of the clinical significance of these observations. No further preprocessing is applied to the text of the reports.

\subsection{Evaluation metrics}
\label{sec:eval_metrics}

\begin{wrapfigure}{r}{0.5\textwidth}
\vspace{-1.cm}
\centerline{\includegraphics[width=0.5\textwidth]{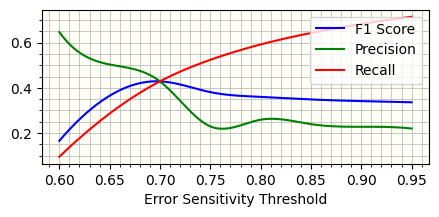}}
\caption{\textbf{Trade-off between precision, recall and F1 score  at various error sensitivity thresholds for error detection module.} The F1 score peaks at a threshold of approximately 0.7, suggesting an optimal balance between precision and recall at this point that we use to evaluate error detection.}
\vspace{-0.3cm}
\label{fig:q_5}
\end{wrapfigure}

In evaluating our image-conditioned autocorrection framework, we use a multi-level approach. For per-token binary classification, we employ standard precision, recall, and F1 scores to gauge individual token accuracy with emphasis on the performance on the minority class—which in our case represents the introduced errors. To adjust the error detection module's performance, we utilize an ``error sensitivity threshold,'' allowing for adjustable sensitivity in error detection. A threshold set closer to 1.0 results in a stringent detection approach, identifying more potential errors, whereas a setting closer to 0.6 yields a more lenient detection module, reducing the likelihood of false 
positives (refer to Fig.~\ref{fig:q_5}). In our experiments, we set the threshold to $0.7$.

\begin{wraptable}{r}{0.58\textwidth}
\vspace{-0.2cm}
\centering
\begin{tabular}{lccc}
\toprule
\textbf{Approach} & \textbf{Precision} & \textbf{Recall} & \textbf{F1-score} \\
\midrule
Patch & \textbf{0.4931} & 0.8763 & \textbf{0.6311} \\
Pool & 0.4862 & 0.8569 & 0.6204 \\
Concatenate & 0.4647 & \textbf{0.9020} & 0.6134 \\
\bottomrule
\end{tabular}
\caption{\textbf{Error Identification Results}. We compare the three approaches for the error detection module introduced in Section~\ref{sec:methods}. This comparison highlights the better performance of the patch approach, outperforming other conditioning methods in per-token classification. We set the ``error sensitivity threshold'' to 0.7 for these results (see Section~\ref{sec:eval_metrics} for details on the threshold).}
\vspace{-0.5cm}
\label{tab:model_evaluation_with_support}
\end{wraptable}

At the report level, we assess overall performance using established natural language generation metrics: BLEU~\citep{miura-etal-2021-improving}, which measures n-gram overlap between generated and reference reports; METEOR~\citep{banerjee-lavie-2005-meteor}, which captures semantic content; and ROUGE-L~\citep{lin-2004-rouge}, focusing on the longest common subsequence in the text. Additionally, we assess the clinical efficacy of the autocorrected reports by utilizing metrics based on the specific pathology classes from the CheXpert dataset ~\citep{irvin2019chexpert}. The results are microaveraged across the 14 classes.

\subsection{Evaluation strategy and baselines}

In our comparative analysis, we assess the performance of a CLIP-based~\citep{radford2021learning} retrieval generative model \citep{pmlr-v158-endo21a} enhanced by our autocorrection framework against previous state-of-the-art models in radiology report generation. These models include R2Gen~\citep{chen-etal-2020-generating}  CMN~\citep{chen2022crossmodal}, PPKED~\citep{10.1609/aaai.v33i01.33018650}, \(\mathcal{M}^2\) TR. PROGRESSIVE~\citep{nooralahzadeh-etal-2021-progressive-transformer}, Contrastive Attention~\citep{liu-etal-2021-contrastive}, AlignTransformer~\citep{you2022aligntransformer}, \(\mathcal{M}^2\) Trans~\citep{cornia2020m2}, ITA~\citep{10.1145/3301275.3302308}, and CvT-212DistilGPT2~\citep{liu2019clinically}. These models, optimized for standard language model loss and augmented with rewards for factual completeness and consistency~\citep{liu-etal-2021-contrastive}, serve as benchmarks for our analysis. Results are cited directly from their respective publications unless otherwise noted. Since, to our knowledge, our work is the first to incorporate an autocorrection framework in radiological report generation models, we do not have direct baselines to compare against. Therefore, in Fig.~\ref{fig:retrieval_example} we also provide a qualitative component that shows the improvements our method offers.

\begin{table*}[ht!]
    \centering
    \resizebox{\textwidth}{!}{\begin{tabular}{lcccccc}
        \toprule
        & \textbf{BLEU-1} & \textbf{BLEU-2} & \textbf{BLEU-3} & \textbf{BLEU-4} & \textbf{METEOR} & \textbf{ROUGE-L F1} \\
        \midrule
        Baseline & 0.238 & 0.201 & 0.155 & 0.091 & 0.139 & 0.227 \\
        \noalign{\vskip1pt}
       \multirow{2}{*}{Patch approach} & \cellcolor{gray!30}\textbf{0.3169} & \cellcolor{gray!30}\textbf{0.2539} 
       & \cellcolor{gray!30}\textbf{0.2091} & \cellcolor{gray!30}\textbf{0.1693} & \cellcolor{gray!30}\textbf{0.2043} & \cellcolor{gray!30}\textbf{0.3544} \\
       \noalign{\vskip-3pt}
        &\cellcolor{gray!30}\textbf{\footnotesize{(0.3073, 0.3266)}} & \cellcolor{gray!30}\textbf{\footnotesize{(0.2429, 0.2648)}} & \cellcolor{gray!30}\textbf{\footnotesize{(0.1971, 0.2211)}} & \cellcolor{gray!30}\textbf{\footnotesize{(0.1568, 0.1817)}} & \cellcolor{gray!30}\textbf{\footnotesize{(0.1806, 0.2280)}} & \cellcolor{gray!30}\textbf{\footnotesize{(0.3388, 0.3700)}} \\
        \noalign{\vskip1pt}
       \multirow{2}{*}{Pool approach} & 0.3056 & 0.2387 & 0.1927 & 0.1523 & 0.1787 & 0.3376 \\
       \noalign{\vskip-3pt}
        & \footnotesize{(0.2961, 0.3152)} & \footnotesize{(0.2280, 0.2494)} & \footnotesize{(0.1811, 0.2044)} & \footnotesize{(0.1401, 0.1646)} & \footnotesize{(0.1560, 0.2015)} & \footnotesize{(0.3231, 0.3522)} \\
        \noalign{\vskip1pt}
        \multirow{2}{*}{Concatenate approach}& 0.3031 & 0.2370 & 0.1915 & 0.1504 & 0.1776 & 0.3369 \\
        \noalign{\vskip-3pt}
        & \footnotesize{(0.2937, 0.3125)} & \footnotesize{(0.2265, 0.2475)} & \footnotesize{(0.1800, 0.2030)} & \footnotesize{(0.1382, 0.1627)} & \footnotesize{(0.1550, 0.2002)} & \footnotesize{(0.3219, 0.3518)} \\
        \bottomrule
    \end{tabular}}
    \caption{\textbf{Corrections Ablation Study}. We report metrics evaluating autocorrected reports against ground truth for different approaches. Baseline indicates the original scores of the uncorrected reports in relation to the groundtruth. Conditioning on patch embeddings outperforms the other conditioning mechanisms on improving the quality of the report. Refer to Section~\ref{sec:methods} for definition of the different approaches. We report 95\% Confidence Intervals in parenthesis.}
    \label{tab:first_eval}
\end{table*}

\begin{table*}[ht!]
    \centering
    \small
    \resizebox{\textwidth}{!}{\begin{tabular}{lclccccc|ccc}
        \toprule
        \multirow{2}{*}{\textbf{Dataset}} & \multirow{2}{*}{\textbf{Method}} & \multicolumn{6}{c|}{NLG Metrics $\uparrow$} & \multicolumn{3}{c}{CE Metrics $\uparrow$} \\
        & & \textbf{BL-1} & \textbf{BL-2} & \textbf{BL-3} & \textbf{BL-4} & \textbf{MTR} & \textbf{RG-L}  & P & R & F1 \\
        \midrule
        \multirow{11}{*}{\rotatebox[origin=c]{90}{MIMIC-CXR}} & Uncorrected retrieval (Ours) & 0.216 &0.108 & 0.056 & 0.029 & 0.051 & 0.187 & 0.192 & 0.173 & 0.183 \\
        & Retrieval with Autocorrection (Ours)  & \cellcolor{gray!30}0.370 & \cellcolor{gray!30}0.234 & \cellcolor{gray!30}0.175 & \cellcolor{gray!30}0.125 & \cellcolor{gray!30}0.112 & \cellcolor{gray!30}0.230 & \cellcolor{gray!30}0.263 & \cellcolor{gray!30}0.444  & \cellcolor{gray!30}0.330\\
        \midrule
        & \textcolor{gray}{R2Gen}\citep{chen-etal-2020-generating} & \textcolor{gray}{0.353} & \textcolor{gray}{0.218} & \textcolor{gray}{0.145} & \textcolor{gray}{0.103} & \textcolor{gray}{0.142} & \textcolor{gray}{0.277} & \textcolor{gray}{0.331} & \textcolor{gray}{0.224} & \textcolor{gray}{0.228} \\
 & \textcolor{gray}{CMN}\citep{chen2022crossmodal} & \textcolor{gray}{0.353} & \textcolor{gray}{0.218} & \textcolor{gray}{0.148} & \textcolor{gray}{0.106} & \textcolor{gray}{0.142} & \textcolor{gray}{0.278} & \textcolor{gray}{0.334} & \textcolor{gray}{0.275} & \textcolor{gray}{0.278} \\
 & \textcolor{gray}{PPKED}\citep{Liu2021ExploringAD} & \textcolor{gray}{0.360} & \textcolor{gray}{0.224} & \textcolor{gray}{0.149} & \textcolor{gray}{0.106} & \textcolor{gray}{0.149} & \textcolor{gray}{0.284} & \textcolor{gray}{-} & \textcolor{gray}{-} & \textcolor{gray}{-} \\
 & \textcolor{gray}{\(\mathcal{M}^2\) TR. PROGRESSIVE}\citep{nooralahzadeh-etal-2021-progressive-transformer} & \textcolor{gray}{0.378} & \textcolor{gray}{0.232} & \textcolor{gray}{0.154} & \textcolor{gray}{0.107} & \textcolor{gray}{0.145} & \textcolor{gray}{0.272} & \textcolor{gray}{0.240} & \textcolor{gray}{0.428} & \textcolor{gray}{0.308} \\
 & \textcolor{gray}{Contrastive Attention}\citep{10.5555/3326943.3327084} & \textcolor{gray}{0.350} & \textcolor{gray}{0.219} & \textcolor{gray}{0.152} & \textcolor{gray}{0.109} & \textcolor{gray}{0.151} & \textcolor{gray}{0.283} & \textcolor{gray}{0.352} & \textcolor{gray}{0.298} & \textcolor{gray}{0.303} \\
 & \textcolor{gray}{AlignTransformer}\citep{10.1007/978-3-030-87199-4_7} & \textcolor{gray}{0.378} & \textcolor{gray}{0.235} & \textcolor{gray}{0.156} & \textcolor{gray}{0.112} & \textcolor{gray}{0.158} & \textcolor{gray}{0.283} & \textcolor{gray}{-} & \textcolor{gray}{-} & \textcolor{gray}{-} \\
 & \textcolor{gray}{ITA}\citep{10.1007/978-3-031-16452-1_54} & \textcolor{gray}{0.395} & \textcolor{gray}{0.253} & \textcolor{gray}{0.170} & \textcolor{gray}{0.121} & \textcolor{gray}{0.147} & \textcolor{gray}{0.284} & \textcolor{gray}{-} & \textcolor{gray}{-} & \textcolor{gray}{-} \\
 & \textcolor{gray}{CvT-212DistilGPT2}\citep{liu2019clinically} & \textcolor{gray}{0.392} & \textcolor{gray}{0.245} & \textcolor{gray}{0.169} & \textcolor{gray}{0.124} & \textcolor{gray}{0.153} & \textcolor{gray}{0.285} & \textcolor{gray}{0.359} & \textcolor{gray}{0.412} & \textcolor{gray}{0.384} \\
 & \textcolor{gray}{RGRG}\citep{Tanida_2023} & \textcolor{gray}{0.373} & \textcolor{gray}{0.249} & \textcolor{gray}{0.175} & \textcolor{gray}{0.126} & \textcolor{gray}{0.168} & \textcolor{gray}{0.264} & \textcolor{gray}{0.461} & \textcolor{gray}{0.475} & \textcolor{gray}{0.447} \\
 \midrule
& \textcolor{black}{RGRG}\citep{Tanida_2023} with autocorrection & \cellcolor{gray!30}{0.375} & \cellcolor{gray!30}{0.243} & \cellcolor{gray!30}{0.151} & \cellcolor{gray!30}{0.116} & \cellcolor{gray!30}{0.157} & \cellcolor{gray!30}{0.271} & \cellcolor{gray!30}\textbf{0.469} & \cellcolor{gray!30}{0.458} & \cellcolor{gray!30}\textbf{0.463} \\

        \bottomrule
    \end{tabular}}
    \caption{\textbf{Quantitative effect of autocorrection on radiology report generation through retrieval}. Evaluation is done against baseline models and an uncorrected retrieval system that was intentionally not optimized to state-of-the-art standards. Note that \textbf{our method was not trained to do report generation}, only error detection and correction. This approach shows the potential of how the performance of a radiology report generative model can be improved through autocorrection of incorrect parts (See performance on RGRG for the effect of autocorrection on a SOTA model). Grayed-out results have been trained explicitly for report generation. We just add them here for completeness. See ~\ref{fig:retrieval_example} for a qualitative example.}
    \label{tab:second_eval}
\vspace{-0.5cm}
\end{table*}

\textbf{Robustness of error correction.} 
Quantitative assessment of our error correction model presents challenges. 
Specifically, standard text generation metrics are limited in conveying \emph{robustness} information. Robustness in the context of error correction models would involve assessing how well the model can handle and correct various types of errors, including those that might not significantly affect n-gram overlap or semantic similarity--for example, errors caused by negations (See Fig.~\ref{fig:attention_maps} for examples). Consider the difference, in a radiological report, between ``no evidence of fracture'' and ``evidence of fracture'': it is only one negation word (``no''), but the meanings are diametrically opposed (See sentences in Fig.~\ref{fig:q_4} and Fig.~\ref{fig:attention_maps}). Traditional NLG metrics might score these two sentences as highly similar because of the high degree of word overlap, despite the fact that the negation leads to a critical error in meaning. To tackle this, we focus on the model's resilience to variations in types and positions of errors introduced into radiological reports that may not be reflected in changes in NLG metrics.

\section{Results and Discussion}

\subsection{Performance Analysis}
\label{sec:analysis}
We present error identification results in Table~\ref{tab:model_evaluation_with_support}. We see that conditioning the error detection module on token embeddings leads to much better performance in error detection than the other conditioning mechanisms highlighted in section~\ref{sec:methods}. We attribute this to the availability of more image contextual information that can be used to improve the classification of the tokens.

\begin{figure*}
\vspace{-0.5cm}
\centerline{\includegraphics[width=1.\textwidth]{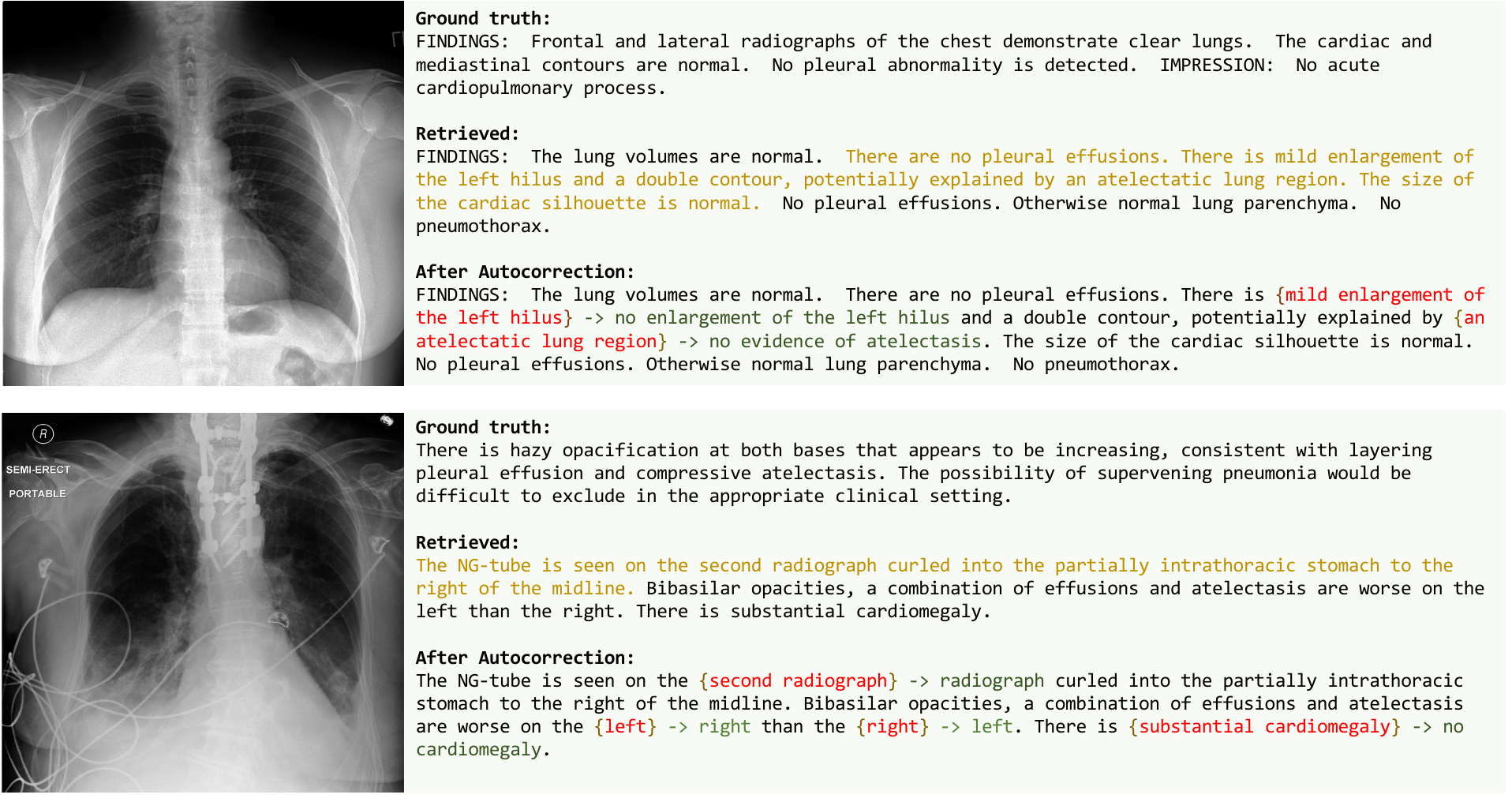}}
\caption{\textbf{Challenges in Autocorrection of Retrieved Reports (See figure~\ref{fig:retrieval_example}):} This figure highlights a significant challenge in autocorrecting retrieval-based radiology reports. It showcases instances where reports, while accurately retrieved, include additional details that are not pertinent to the current diagnosis (highlighted in yellow). These extraneous segments, although potentially relevant, deviate from the ground truth and are not efficiently identified or removed by the autocorrect system. This limitation underscores the need for enhanced discernment capabilities in the framework to differentiate between essential and non-essential information in the context of clinical diagnosis. Beyond that, autocorrection happens just as it would as illustrated in Figure~\ref{fig:attention_maps}}
\label{fig:retrieval_2}
\vspace{-0.5cm}
\end{figure*}

We also show the error correction results  in Table~\ref{tab:first_eval}. The corrected reports are compared against the ground truth along with the original incorrect reports whose scores are listed as baselines. Error correction is only as good as the error identification module hence the observed better performance of the approach that involves conditioning on the token embeddings in Table~\ref{tab:first_eval}. Overall, as observed in Table~\ref{tab:second_eval} our proposed framework demonstrates an ability to significantly improve the quality of reports generated by a retrieval model almost doubling the F1 score on clinical efficacy along with significant improvement in the NLG metrics.

\subsection{Error Correction Impact}
The system's error correction capabilities have a significant impact on correcting misidentification of anatomical location misidentification and severity of findings examples shown in Fig.s~\ref{fig:q_4}, \ref{fig:retrieval_example}, \ref{fig:attention_maps} as well as the inappropriate use of terminologies. The correction of errors related to location misindenfitication and severity of findings is indeed clinically relevant as they can as they directly influence the diagnostic interpretations and subsequent patient management strategies. For example, in Fig.~\ref{fig:attention_maps} the autocorrection framework corrected orientation of a patient using visual cues from shoulders and head regions as well reconciling discrepancies between the described radiographic view and the observed anatomical landmarks—turning `lateral radiographs' into `frontal radiographs' and addressing `opacification in the lungs.' Nonetheless, there is a challenge in correcting errors related to incorrect predictions or omissions of findings as these scenarios require a level of clinical inference akin to report generation, which falls outside the intended scope of our autocorrection framework (See Figs.~\ref{fig:retrieval_example} and~\ref{fig:attention_maps}). Only errors that have been flagged as incorrect can be corrected, meaning the errors have to be present in the first place. This could be an area of focus for future improvements.

\subsection{Guardrail for automatic report generation}
\label{sec:retrieval}
We also qualitatively evaluate the potential of autocorrection in improving radiological report generation methods. This system was layered atop a retrieval-based report generation model, which was not optimized for SOTA results. However, the integration of autocorrection demonstrated a noteworthy enhancement in the accuracy and relevance of generated reports (see Table~\ref{tab:first_eval}).

\textbf{Case Analysis:} Consider Fig.~\ref{fig:retrieval_example}, where the ground truth report noted a ``focal consolidation in the left lung base, mild vascular engorgement, bilateral apical pleural thickening, aortic arch calcifications, and a heart size that is top normal.'' In contrast, the retrieved report initially presented an inaccurately ``stable and severely enlarged heart, along with a clear lung field,'' diverging significantly from the ground truth as seen in the retrieved report.
 The autocorrection system modified key descriptors in the retrieved report, aligning them closer to the ground truth. For example, it corrected ``well expanded'' to ``adequately aerated'' and ``stably and severely enlarged'' to ``normal in size,'' among other adjustments. This indicates the system's potential in identifying and rectifying specific inaccuracies in radiological reports.


When benchmarked autocorrection layered atop of retrieval against established methods like R2Gen, CMN, PPKED (see Table~\ref{tab:second_eval}), our autocorrection approach displayed commendable performance. It closely approached the baseline method and showed its utility in enhancing the accuracy of diagnostic reporting in radiology (refer to Table~\ref{tab:second_eval} and Fig.~\ref{fig:retrieval_example}). This suggests that even non-SOTA retrieval-based systems can be significantly improved with effective autocorrection mechanisms. The ability to adjust key clinical descriptors accurately is crucial, as these directly impact the clinical interpretation and subsequent patient management decisions.

\section{Conclusion}
\label{sec:conclusion}

In this study, we have introduced a framework for medical report autocorrection with a strong emphasis on error detection and correction within radiological texts grounded in the images they describe. Our model leverages image-conditioned data processing to enhance the accuracy of medical report generation. This approach not only ensures the generation of clinically coherent reports but also contributes to the reduction of errors that could significantly impact patient care.

\textbf{Ethical Considerations.} The domain of automated medical text correction holds immense potential to enhance healthcare delivery. Nevertheless, the implications of incorrect corrections are substantial, with the possibility of negatively affecting patient outcomes. While our system represents a step towards minimizing human error, it also raises concerns regarding overdependence on automation \citep{agarwal2023combining}. While autonomous reporting may be a distant possibility \citep{saenz2023autonomous}, we expect that systems may be first employed as decision support tools with guardrails \citep{sanchez2023ai} to ensure that clinicians remain engaged and vigilant in the decision-making process.

\section{Limitations}
\label{sec:limitations}


Given the high stakes of medical autocorrection, especially within the context of clinical data, discussing limitations here is key. The primary dataset used in this paper may not encompass the complete spectrum of errors inherent in radiological reports. Given the variability of medical data, including rare but critical scenarios, it remains a significant challenge to capture every possible error type within the training corpus. This limitation could affect the model’s ability to generalize to errors not represented in the dataset. 

Furthermore, the proposed framework operates on the assumption that input reports are largely accurate, with only a few factual errors. This assumption might hold true for real-world reports written by radiologists who are not likely to make errors. However, for reports that are poorly constructed or contain multiple, compounded errors the framework will perform poorly and will amount to the solving of a report generation task. The model’s performance is contingent upon this assumption, which may not always be a given in clinical settings.

\textbf{Generalization Issues.} The model has been trained and validated on a dataset which may not cover the full diversity of medical language and errors. Consequently, there is a risk that the model may underperform when faced with novel errors or those that manifest in contexts divergent from the training data. To this end, we propose a strategy for real-world clinical applications——using the same approach, tailor error injection based on specific hospital environments before error correction is learnt for that setting.

\textbf{Future Work.} Addressing these limitations presents clear avenues for future work. Tools that identify clinical entities and relationships in the report \citep{jain2021radgraph, khanna2023radgraph2} could also be incorporated to guide the correction by separating style from content \citep{yan2023style}. Classification of severity of errors, using schemes as introduced in~\citep{jeong2023multimodal}, could allow for a more clinically meaningful analysis of the error reduction. Enriching the dataset with a broader range of error types, investigating context-aware error detection mechanisms, and developing robust models that can handle a variety of report qualities are critical next steps. 



{
    \small
    \bibliography{main}
}

\end{document}